\documentclass[accepted]{uai2021} 

\usepackage[american]{babel}

\usepackage{natbib} 
\usepackage{mathtools} 
\usepackage{booktabs} 
\usepackage{tikz} 
\usepackage{amsmath}
\usepackage{amsfonts}
\usepackage{float}
\usepackage{bashful}
\usepackage{pdfpages}
\usepackage{xcolor}
\usepackage{graphicx}
\usepackage[caption = false]{subfig}

\usepackage{xr-hyper}
\usepackage{hyperref}
\makeatletter
\newcommand*{\addFileDependency}[1]{
  \typeout{(#1)}
  \@addtofilelist{#1}
  \IfFileExists{#1}{}{\typeout{No file #1.}}
}
\makeatother

\newcommand*{\myexternaldocument}[1]{%
    \externaldocument{#1}%
    \addFileDependency{#1.tex}%
    \addFileDependency{#1.aux}%
}

\myexternaldocument{appendix}

\title{Empirical Frequentist Coverage\\of Deep Learning Uncertainty Quantification Procedures}

\author[1]{\href{mailto:Benjamin Kompa <benjamin_kompa@hms.harvard.edu>?Subject=Your UAI 2021 paper}{Benjamin~Kompa}{}}
\author[2]{Jasper~Snoek}
\author[3]{Andrew~Beam}
\affil[1]{%
    Department of Biomedical Informatics\\
    Harvard Medical School
}
\affil[2]{%
    Google Research
}
\affil[3]{
Department of Epidemiology\\
 Harvard School of Public Health
}

\begin{document}
\maketitle

\begin{abstract}
Uncertainty quantification for complex deep learning models is increasingly important as these techniques see growing use in high-stakes, real-world settings. Currently, the quality of a model's uncertainty is evaluated using point-prediction metrics such as the negative log-likelihood (NLL), expected calibration error (ECE) or the Brier score on heldout data. Marginal coverage of prediction intervals or sets, a well-known concept in the statistical literature, is an intuitive alternative to these metrics but has yet to be systematically studied for this class of models. With marginal coverage, and the complementary notion of the width of a prediction interval, downstream users of a deployed machine learning models can better understand uncertainty quantification both on a global dataset level and on a per-sample basis. In this study, we provide the first large scale evaluation of the empirical frequentist coverage properties of well known uncertainty quantification techniques on a suite of regression and classification tasks. We find that, in general, some methods do achieve desirable coverage properties on \emph{in distribution} samples, but that coverage is not maintained on out-of-distribution data. Our results demonstrate the failings of current uncertainty quantification techniques as dataset shift increases and reinforce coverage as an important metric in developing models for real-world applications.
\end{abstract}

\section{Introduction}\label{sec:intro}
Predictive models based on deep learning have seen dramatic improvement in recent years \citep{lecun2015deep}, which has led to widespread adoption in many areas. For critical, high-stakes domains such as medicine or self-driving cars, it is imperative that mechanisms are in place to ensure safe and reliable operation. Crucial to the notion of safe and reliable deep learning is the effective quantification and communication of \emph{predictive uncertainty} to potential end-users of a system. In medicine, for instance, understanding predictive uncertainty could lead to better decision-making through improved allocation hospital resources, detecting dataset shift in deployed algorithms, or helping machine learning models abstain from making a prediction \citep{Kompa2021-lf}. For medical classification problems involving many possible labels (i.e. creating a \emph{differential diagnosis}), methods that provide a set of possible diagnoses when uncertain are natural to consider and and align more closely with the differential diagnosis procedure used by physicians. The prediction sets and intervals we propose in this work are a new way to quantify uncertainty in machine learning models and provide intuitive metrics for downstream users such as clinicians. 

Many approaches have recently been proposed to quantify uncertainty and generally fall into two broad categories: ensembles and approximate Bayesian methods. Deep ensembles \citep{Lakshminarayanan2017-pv} aggregate information from multiple individual models to provide a measure of uncertainty that reflects the ensembles' agreement about a given data point. Bayesian methods offer direct access to predictive uncertainty through the posterior predictive distribution, which combines prior knowledge with the observed data. Although conceptually elegant, calculating exact posteriors of even simple neural models is computationally intractable~\citep{Yao2019-ia, Neal1996-li}, and many approximations have been developed \citep{Hernandez-Lobato2015-fo, Blundell2015-ms, Graves2011-er, Pawlowski2017-si, Hernandez-Lobato2015-un, Louizos2016-ki, Louizos2017-wo}. Though approximate Bayesian methods scale to modern sized data and models, recent work has questioned the quality of the uncertainty provided by these approximations \citep{Yao2019-ia,Wenzel2020-ui, Ovadia2019-tt}.

Previous work assessing the quality of uncertainty estimates have focused on calibration metrics and scoring rules such as the negative-loglikelihood (NLL), expected calibration error (ECE), and Brier score. Here we provide a complementary perspective based on the notion of empirical \emph{coverage}, a well-established concept in the statistical literature \citep{wasserman2013all} that evaluates the quality of a predictive \emph{set} or \emph{interval} instead of a point prediction. Informally, coverage asks the question: If a model produces a predictive uncertainty interval, how often does that interval actually contain the observed value? Ideally, predictions on examples for which a model is uncertain would produce larger intervals and thus be more likely to cover the observed value. 

In this work, we focus on marginal coverage \textit{over a dataset} $\mathcal{D}'$ for the canonical $\alpha$ value of $0.05$, i.e. 95\% prediciton intervals. For a machine learning model that produces a 95\% prediction interval $\hat{\mathcal{C}_n}({x_n})$ based on the training dataset $\mathcal{D}$, we consider what fraction of the points in the dataset $\mathcal{D}'$ have their true label contained in $\hat{\mathcal{C}_n}({x_{n+1}})$ for ${x_{n+1}} \in \mathcal{D}'$. To measure the robustness of these intervals, we also consider cases when the generating distributions for $\mathcal{D}$ and $\mathcal{D}'$ are not the same (i.e. dataset shift). 

\begin{figure}[h]
    \centering
    \includegraphics[width=.5\textwidth]{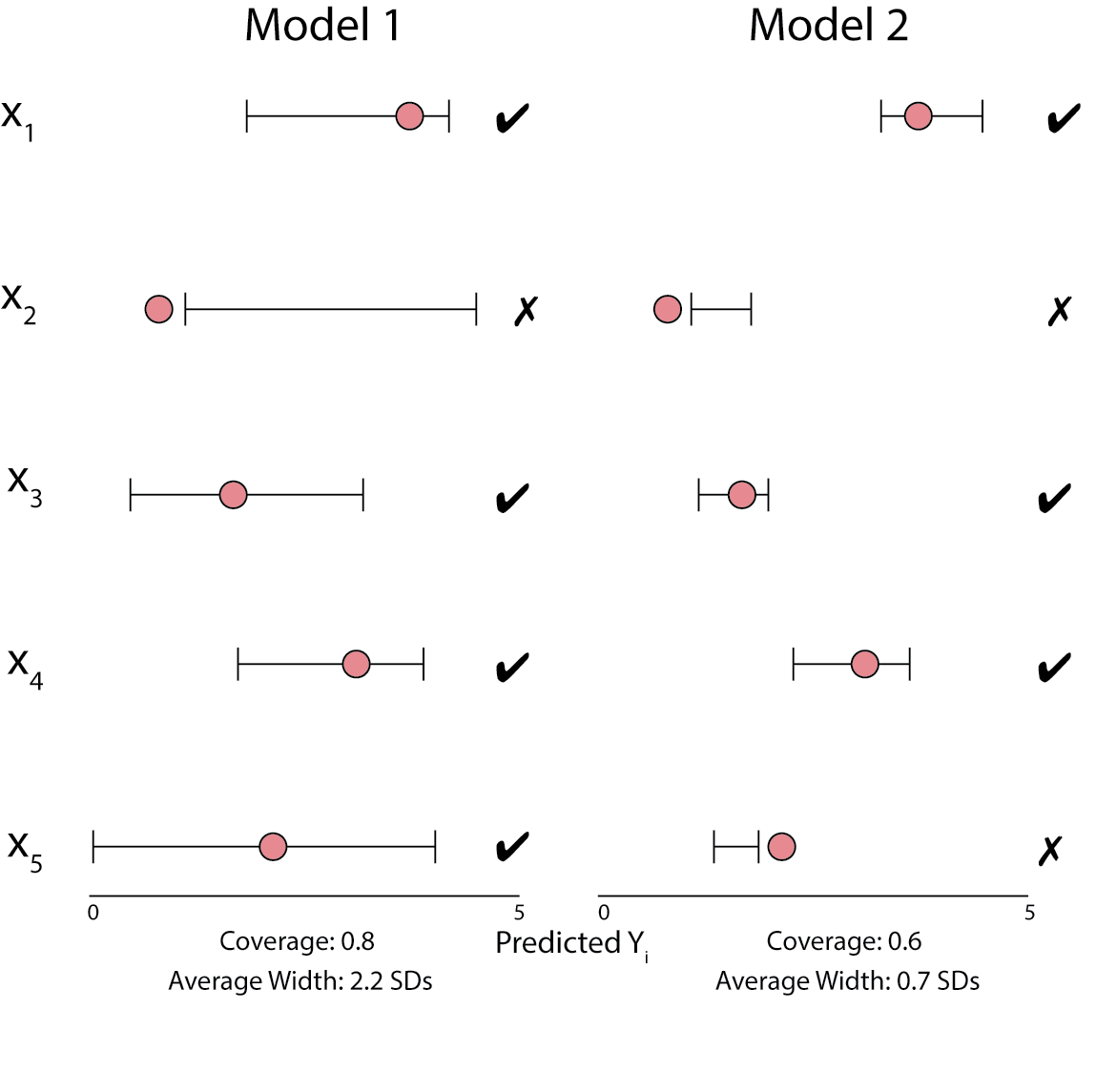}
    \caption{An example of the coverage properties for two methods of uncertainty quantification. In this scenario, each model produces an uncertainty interval for each $x_i$ which attempts to cover the true $y_i$, represented by the red points. Coverage is calculated as the fraction of true values contained in these regions, while the width of these regions is reported in terms of multiples of the standard deviation of the training set $y_i$ values.}
    \label{fig:coverage}
\end{figure}

Figure \ref{fig:coverage} provides a visual depiction of marginal coverage over a dataset for two hypothetical regression models. Throughout this work, we refer to ``marginal coverage over a dataset'' as ``coverage''. 

For a machine learning model that produces predictive uncertainty estimates (i.e. approximate Bayesian methods and ensembling), coverage encompasses both the aleatoric and epistemic uncertainties \citep{Gal2016-yd} produced by these models. The predictions from these models can be written as:

\begin{equation}
    \hat{y} = f(x) + \epsilon
\end{equation}
where epistemic uncertainty is captured in the $f(x)$ component while aleatoric uncertainty is considered in the $\epsilon$ term. Since coverage captures how often the predicted interval of $\hat{y}$ contains the true value, it captures the contributions from both types of uncertainty. 

A complementary metric to coverage is \emph{width}, which is the size of the prediction interval or set. In regression problems, we typically measure width in terms of the standard deviation of the true label in the training set. As an example, a prediction interval could have 90\% marginal coverage with an average width of 2 standard deviations.  For classification problems, width is simply the average size of a prediction set. Width can provide a relative ranking of different methods, i.e. given two methods with the same level of coverage we should prefer the method that provides intervals with smaller widths. 

\textbf{Contributions:} In this study we investigate the empirical  coverage properties of prediction intervals constructed from a catalog of popular uncertainty quantification techniques such as ensembling, Monte-Carlo dropout, Gaussian processes, and stochastic variational inference.  We assess the coverage properties of these methods on nine regression tasks and two classification tasks with and without dataset shift. These tasks help us make the following contributions: 

\begin{itemize}
    \item We introduce coverage and width over a dataset as natural and interpretable metrics for evaluating predictive uncertainty for deep learning models.
    \item A comprehensive set of coverage evaluations on a suite of popular uncertainty quantification techniques.
    \item An examination of how dataset shift affects these coverage properties.
\end{itemize}
\begin{figure*}[ht]
    \centering
    \includegraphics[width=\textwidth]{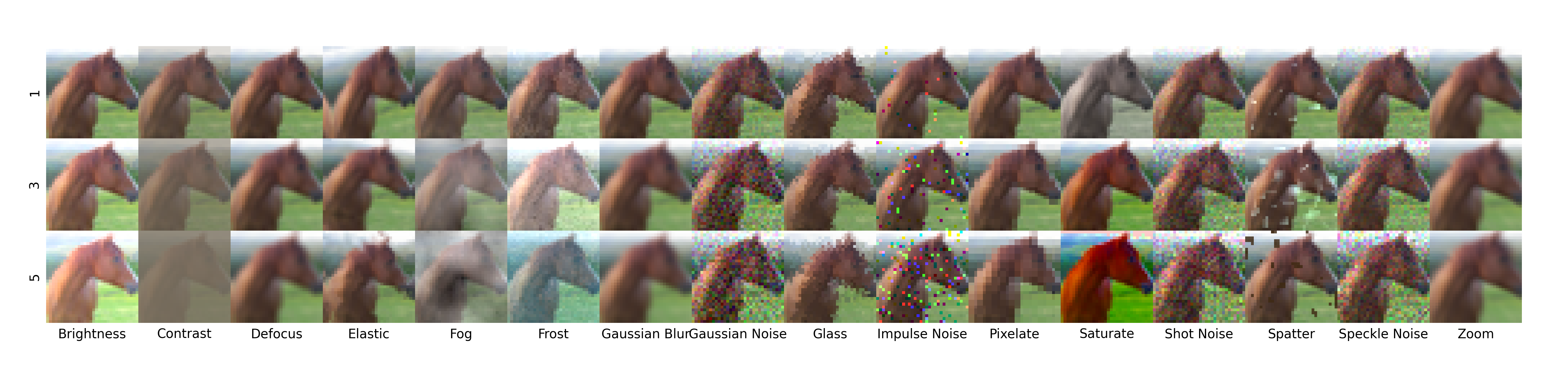}
    \caption{An example of the corruptions in CIFAR-10-C from \cite{Hendrycks2016-ov}. The 16 different corruptions have 5 discrete levels of shift, of which 3 are shown here. The same corruptions were applied to ImageNet to form the ImageNet-C dataset.}
    \label{fig:cifar10c}
\end{figure*}
\section{Background and Related Work}
\textbf{Obtaining Predictive Uncertainty Estimates}\\ Several lines of work focus on improving approximations of the posterior of a Bayesian neural network \citep{Graves2011-er, Hernandez-Lobato2015-fo, Blundell2015-ms, Hernandez-Lobato2015-un, Louizos2016-ki,  Pawlowski2017-si,  Louizos2017-wo}. \citet{Yao2019-ia} provide a comparison of many of these methods and highlight issues with common metrics of comparison, such as test-set log likelihood and RMSE. Good scores on these metrics often indicates that the model posterior happens to match the test data rather than the true posterior \citep{Yao2019-ia}. \cite{Maddox2019-en} developed a technique to sample the approximate posterior from the first moment of SGD iterates. \citet{Wenzel2020-ui} demonstrated that despite advances in these approximations, there are still outstanding challenges  with Bayesian modeling for deep networks.

Alternative methods that do not rely on estimating a posterior over the weights of a model can also be used to provide uncertainty estimates. \citet{Gal2016-yd}, for instance, demonstrated that Monte Carlo dropout is related to a variational approximation to the Bayesian posterior implied by the dropout procedure. \citet{Lakshminarayanan2017-pv} used ensembling of several neural networks to obtain uncertainty estimates. \citet{Guo2017-ly} established that temperature scaling provides well calibrated predictions on an i.i.d test set. More recently, \citet{Van_Amersfoort2020-ho} showed that the distance from the centroids in a RBF neural network yields high quality uncertainty estimates. \cite{Liu2020-ig} also leveraged the notion of distance  (in the form of an approximate Gaussian process covariance function) to obtain uncertainty estimates with their Spectral-normalized Neural Gaussian Processes.

\textbf{Assessments of Uncertainty Properties under Dataset Shift}\\ \citet{Ovadia2019-tt} analyzed the effect of dataset shift on the accuracy and calibration of Bayesian deep learning methods. Their large scale empirical study assessed these methods on standard datasets such as MNIST, CIFAR-10, ImageNet, and other non-image based datasets. Additionally, they used translations, rotations, and corruptions \citep{Hendrycks2016-ov} of these datasets to quantify performance under dataset shift. They found stochastic variational inference (SVI) to be promising on simpler datasets such as MNIST and CIFAR-10, but more difficult to train on larger datasets. Deep ensembles had the most robust response to dataset shift.

\textbf{Definitions of Coverage}\\ 
Given features ${x_i} \in \mathbb{R}^d$ and a response $y_i \in \mathbb{R}$ for some dataset $\mathcal{D} = \{({x_i},y_i)\}_{i=1}^n$, \cite{Barber2019-ra} define \textit{distribution-free} marginal coverage in terms of a set $\hat{\mathcal{C}}_n({x})$ and a level $\alpha \in [0,1]$.  The set $\hat{\mathcal{C}}_n({x})$ is said to have coverage at the $1-\alpha$ level if for all distributions $P \in \mathbb{R}^d \times \mathbb{R}$ where $({x},y) \sim P$, the following inequality holds:
\begin{equation}
\label{eq:coverage}
    \mathbb{P}\{y_{n+1} \in \hat{\mathcal{C}}_n({x_{n+1}})\} \geq 1-\alpha
\end{equation}
For new samples beyond the first $n$ samples in the training data, there is a $1-\alpha$ probability of the true label of the test point being contained in the set $\hat{\mathcal{C}}_n({x_{n+1}})$. This set can be constructed using a variety of procedures. For example, in the case of simple linear regression a prediction interval for a new point $x_{n+1}$ can be constructed\footnote{A well-known result from the statistics literature (c.f. chapter 13 of \citet{wasserman2013all}) is that the interval is given by $\hat{y}_{n+1} \pm t_{n-2}s_y\sqrt{1/n + (x_{n+1}-\bar{x})^2/((n-1)s_x^2)}$, where $\hat{y}_{n+1}$ is the predicted value, $t_{n-2}$ is the $1-\alpha/2$ critical value from a t-distribution with $n-2$ degrees of freedom, $\bar{x}$ is the mean of $x$ in the training data, and $s_y, s_x$ are the standard deviations for $y$ and $x$ respectively. such that (\ref{eq:coverage}) holds asymptotically. However, for more complicated models, closed form solutions with coverage guarantees are unavailable, and constructing these intervals via the bootstrap \citep{efron1982jackknife}) can be computationally infeasible or fail to provide the correct coverage \citep{chatterjee2011bootstrapping}.} using a simple, closed-form solution. 

An important and often overlooked distinction is that of marginal and conditional coverage. In conditional coverage, one considers 
\begin{equation}
\label{eq:conditional-coverage}
    \mathbb{P}\{y_{n+1} \in \hat{\mathcal{C}}_n({x_{n+1}})|{x_{n+1}=x}\} \geq 1-\alpha
\end{equation}
The probability has been conditioned on specific features. This is potentially a more useful version of coverage to consider because one could make claims for specific instances rather than over the broader distribution $P$. However, it is impossible in general to have conditional coverage guarantees \citep{Barber2019-ra}.

Another important point to consider is that while the notion of a confidence interval may seem natural to consider in our analysis, confidence intervals estimate global statistics over repeated trials of data and generally come with guarantees about how often these statistics lie in said intervals. In our study, this is not the case. Although we estimate coverage across many datasets, we are not aiming to estimate an unknown statistic of the data. We would like understand the empirical coverage properties of machine learning models.

\section{Methods}
For features ${x_i} \in \mathbb{R}^d$ and a response $y_i \in \mathbb{R}$ for some dataset $\mathcal{D} = \{({x_i},y_i)\}_{i=1}^n$, we consider the prediction intervals or sets $\hat{\mathcal{C}}_n({x})$ in regression and classification settings, respectively. Unlike in the definitions of marginal and conditional coverage, we do not assume that $(x,y)\sim P$ always holds true. Thus, we consider the marginal coverage on a dataset $\mathcal{D}'$, for some new test set that may have undergone dataset shift from the generating distribution of the training set $\mathcal{D}$.

In both the regression and classification settings, we analyzed the coverage properties of prediction intervals and sets of five different approximate Bayesian and non-Bayesian approaches for uncertainty quantification. These include Dropout \citep{Gal2016-yd, Srivastava2015-ow}, ensembles \citep{Lakshminarayanan2017-pv}, Stochastic Variational Inference \citep{Blundell2015-ms, Graves2011-er, Louizos2016-ki, Louizos2017-wo, Wen2018-fl}, and last layer approximations of SVI and Dropout \citep{Riquelme2018-kg}. Additionally, we considered prediction intervals from linear regression and the 95\% credible interval of a Gaussian process with the squared exponential kernel as baselines in regression tasks. For classification, we also considered temperature scaling \citep{Guo2017-ly} and the softmax output of vanilla deep networks~\citep{Hendrycks2016-ov}.

\subsection{Regression Methods and Metrics}\label{sec:regression-methods}

We evaluated the coverage properties of these methods on nine large real world regression datasets used as a benchmark in \citet{Hernandez-Lobato2015-fo} and later~\citet{Gal2016-yd}. We used the training, validation, and testing splits publicly available from \cite{Gal2016-yd} and performed nested cross validation to find hyperparameters. On the training sets, we did 100 trials of a random search over hyperparameter space of a multi-layer-perceptron architecture with an Adam optimizer \citep{Kingma2014-hf} and selected hyperparameters based on RMSE on the validation set. 

Each approach required slightly different ways to obtain a 95\% prediction interval. For an ensemble of neural networks, we trained $N=40$ vanilla networks and used the 2.5\% and 97.5\% quantiles as the boundaries of the prediction interval. For dropout and last layer dropout, we made 200 predictions per sample and similarly discarded the top and bottom 2.5\% quantiles. For SVI, last layer SVI (LL SVI), and Gaussian processes we had approximate variances available for the posterior which we used to calculate the prediction interval. We calculated 95\% prediction intervals from linear regression using the closed-form solution. 

Then we calculated two metrics: 

\begin{itemize}
    \item \textbf{Coverage}: A sample is considered covered if the true label is contained in this 95\% prediction interval. We average over all samples in a test set to estimate a method's marginal coverage on this dataset.
    \item \textbf{Width}: The width is the average over the test set of the ranges of the 95\% prediction intervals. 
\end{itemize}

Coverage measures how often the true label is in the prediction region while width measures how specific that prediction region is. Ideally, we would have high levels of coverage with low levels of width on in-distribution data. As data becomes increasingly out of distribution, we would like coverage to remain high while width increases to indicate model uncertainty. 

\subsection{Classification Methods and Metrics} \label{sec:class-methods}
\citet{Ovadia2019-tt} evaluated model uncertainty on a variety of datasets publicly available. These predictions were made with the five apprxoimate Bayesian methods describe above, plus vanilla neural networks, with and without temperature scaling. We focus on the predictions from MNIST, CIFAR-10, CIFAR-10-C, ImageNet, and ImageNet-C datasets. For MNIST, we calculated coverage and width of model prediction intervals on rotated and translated versions of the test set. For CIFAR-10, \cite{Ovadia2019-tt} measured model predictions on translated and corrupted versions of the test set from CIFAR-10-C \citep{Hendrycks2016-ov}. For ImageNet, we only considered the coverage and width of prediction sets on the corrupted images of ImageNet-C \citep{Hendrycks2016-ov}. Each of these transformations (rotation, translation, or any of the 16 corruptions) has multiple levels of shift. Rotations range from 15 to 180 degrees in 15 degrees increments. Translations shift images every 2 and 4 pixels for MNIST and CIFAR-10, respectively. Corruptions have 5 increasing levels of intensity. Figure \ref{fig:cifar10c} shows the effects of the 16 corruptions in CIFAR-10-C at the first, third, and fifth levels of intensity.

\begin{figure}[h]
    \centering
    \includegraphics[width=.5\textwidth]{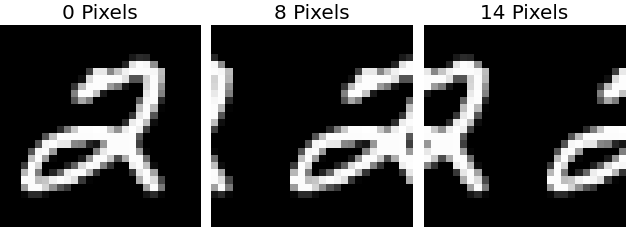}
    \caption{Several examples of the ``rolling'' translation shift that moves an image across an axis.}
    \label{fig:rolling}
\end{figure}

Given $\alpha \in (0,1)$, the $1-\alpha$ prediction set $\mathcal{S}$ for a sample ${x_i}$ is the minimum sized set of classes such that 
\begin{equation}
    \sum_{c \in \mathcal{S}} p(y_c|{x_i}) \geq 1-\alpha
\end{equation}

This consists of the top $k_i$ probabilities such that $1-\alpha$ probability has been accumulated. This inherently assumes that the label are unordered categorical classes such that including classes $1$ and $K$ does not imply that all classes between are also included in the set $\mathcal{S}$. Then we can define: 
\begin{itemize}
    \item \textbf{Coverage:} For each dataset point, we calculate the $1-\alpha$ prediction set of the label probabilities, then coverage is what fraction of these prediction sets contain the true label. 
    \item \textbf{Width:} The width of a prediction set is simply the number of labels in the set, $|\mathcal{S}|$. We report the average width of prediction sets over a dataset in our figures. 
\end{itemize}
Although both calibration \citep{Guo2017-ly} and coverage can involve a probability over a model's output, calibration only considers the most likely label and it's corresponding probability, while coverage considers the the top-$k_i$ probabilities. In the classification setting, coverage is more robust to label errors as it does not penalize models for putting probability on similar classes.

\section{Results}
\subsection{Regression}
\begin{figure}[h]
\subfloat[Coverage and width across regression datasets]{\includegraphics[width = .5\textwidth]{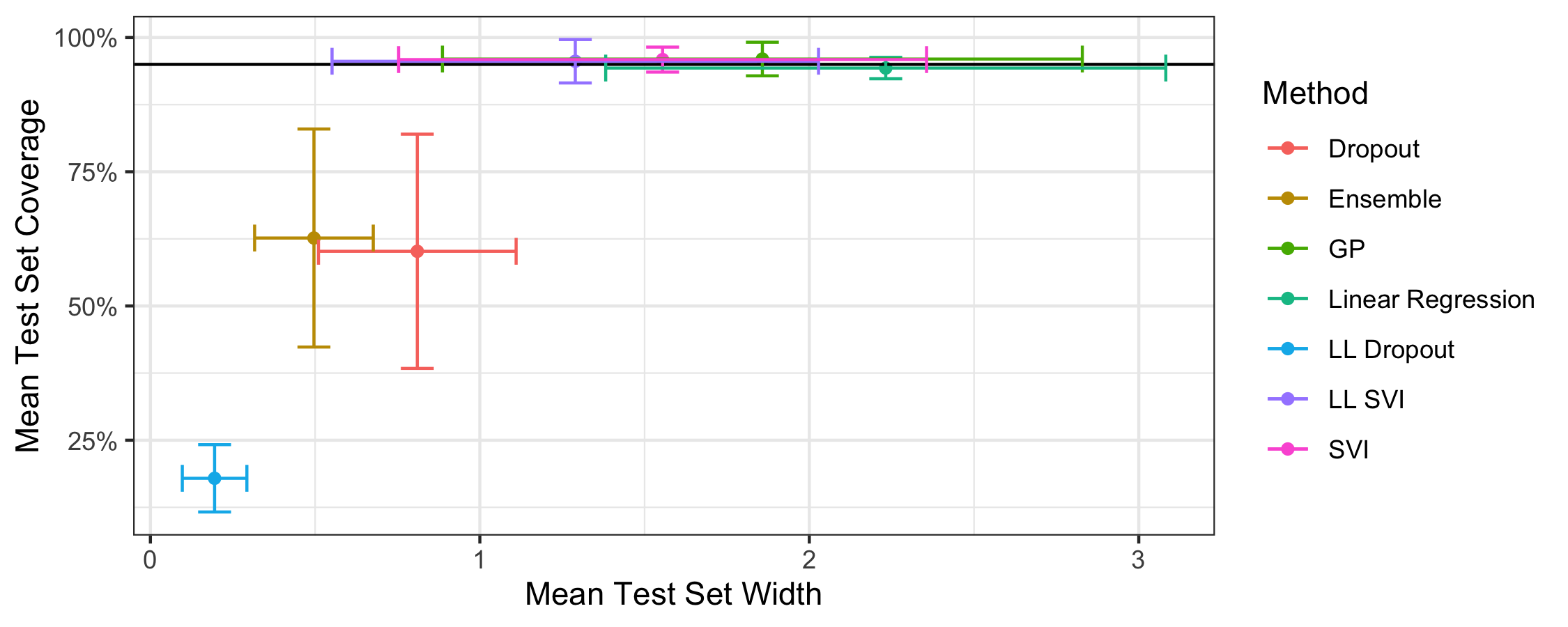}} \\
\subfloat[Detailed view]{\includegraphics[width = .5\textwidth]{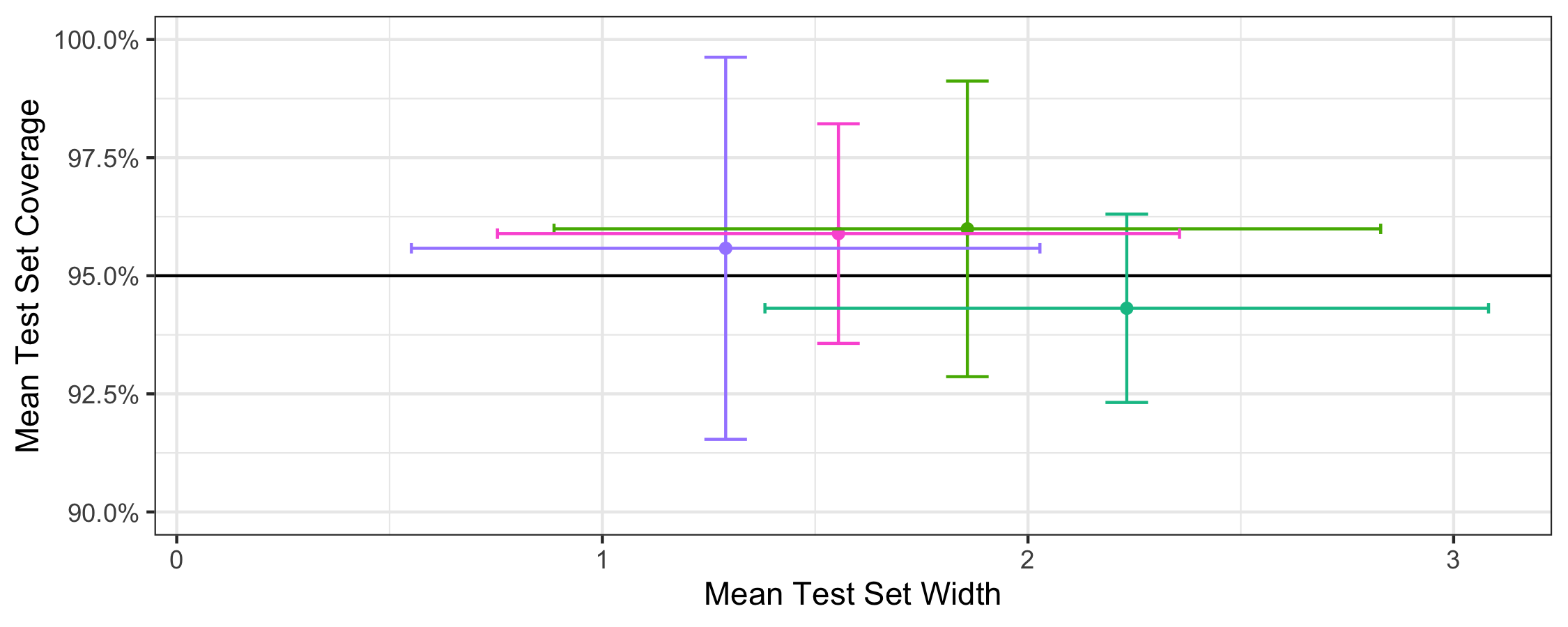}}
\caption{The mean coverage and widths of models' prediction intervals average over the nine regression datasets we considered (\textbf{panel a}). Error bars indicate the standard deviation for both coverage and width across all experiements. In \textbf{panel b} we observe that the four methods which maintained 95\% coverage did so because they had appropriately wide prediction intervals. LL SVI had the lowest average width while maintaining at least 95\% coverage.}
\label{fig:reg_cov_width}
\end{figure}

Figure \ref{fig:reg_cov_width} plots the mean test set coverage and width for the regression methods we considered averaged over the nine regression datasets. Error bars demonstrate that for low performing methods such as ensembling, dropout, and LL dropout, there is high variability in coverage levels and widths across the datasets.

We observe that several methods perform well across the nine datasets. In particular, LL SVI, SVI, and GPs all exceed the 95\% coverage threshold on average, and linear regression comes within statistical sampling error of this threshold. Over the regression datasets we considered, LL SVI had the lowest mean width while maintaining at least 95\% coverage. For specific values of coverage and width for methods on a particular dataset, see Tables \ref{tab:regression-coverage} and \ref{tab:regression-width} in the appendix. 

Figure \ref{fig:reg_cov_width} also demonstrates an important point that will persist through our results. Coverage and width are directly related. Although high coverage can and ideally does occur when width is low, we typically observe that high levels of coverage occur in conjunction with high levels of width.

\subsection{MNIST}
In the classification setting, we begin by calculating coverage and width for predictions from \citet{Ovadia2019-tt} on MNIST and shifted MNIST data. \citet{Ovadia2019-tt} used a LeNet architecture and we refer to their manuscript for more details on their implementation.

\begin{figure}[]
\centering
\includegraphics[width = .5\textwidth]{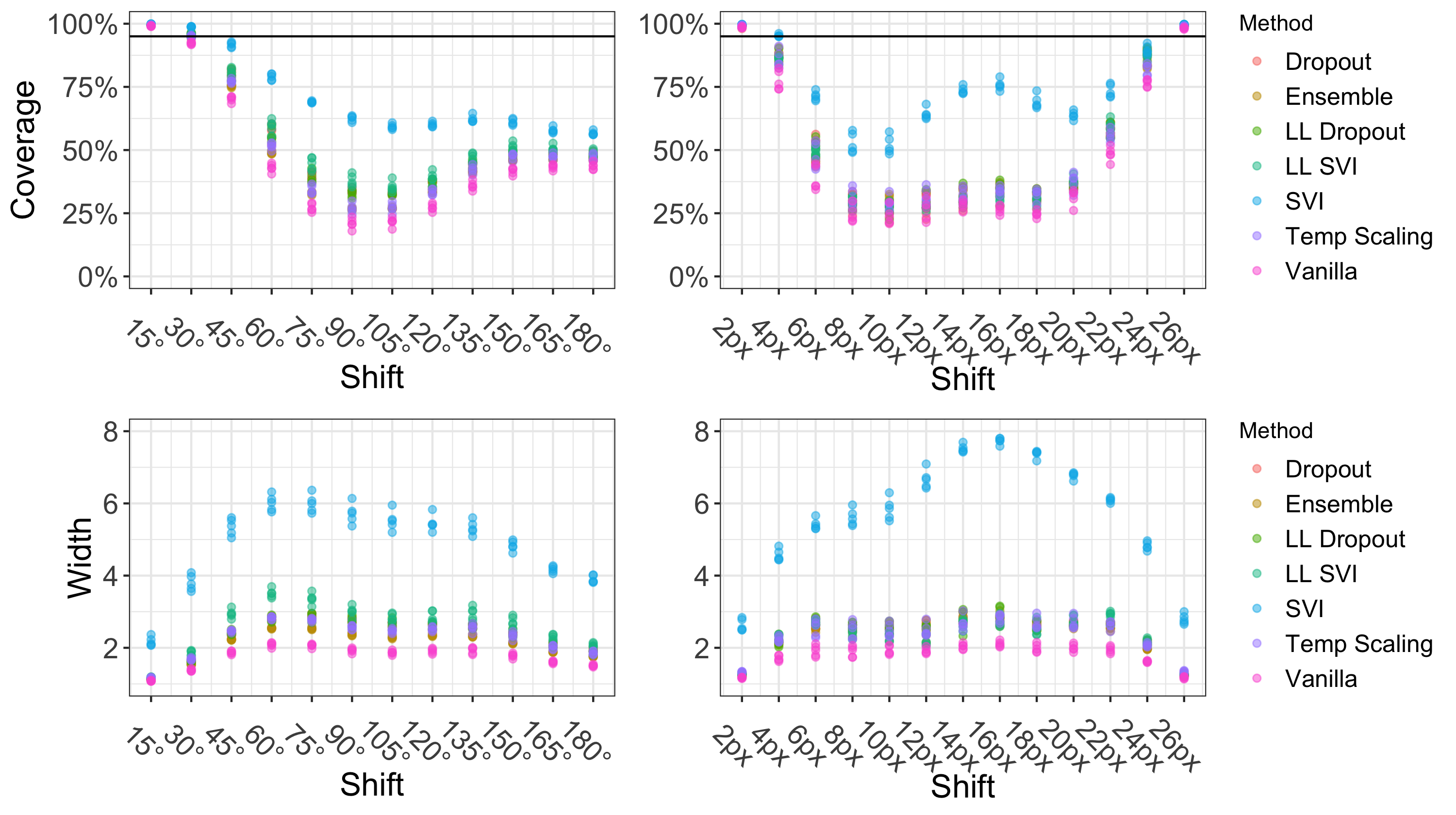}
\caption{The effect of rotation  and translation on coverage and width, respectively, for MNIST.}
\label{fig:mnist}
\end{figure}

Figure \ref{fig:mnist} shows how coverage and width co-vary as dataset shift increases. The elevated width for SVI on these dataset splits indicate that the posterior predictions of label probabilities were the most diffuse to begin with among all models. In Figure \ref{fig:mnist}, all seven models have at least 95\% coverage with a 15 degree rotation shift. Most models don't see an appreciable increase in the average width of the 95\% prediction set, except for SVI. The average width for SVI jumps to over 2 at 15 degrees rotation. As the amount of shift increases, coverage decreases across all methods in a comparable way. In the rotation shifts, we observe that coverage increases and width decreases after about 120 degrees of shift. This is likely due to some of the natural symmetry of several digits (i.e. 0 and 8 look identical after 180 degrees of rotation). 

SVI maintains higher levels of coverage, but with a compensatory increase in width. In fact, there is a Pearson correlation of 0.9 between the width of the SVI prediction set and the distance from the maximum shift of 14 pixels. The maximum shift occurs when the original center of the image is broken across the edge as the image rolls to the right. Figure \ref{fig:rolling}'s right most example is a case of the maximum shift of 14 pixels on a MNIST digit. This strong correlation between width and severity of shift for some methods makes the width of a prediction set at a fixed $\alpha$ level a natural proxy to detect dataset shift. For this simple dataset, SVI outperforms other models with regards to coverage and width properties. It is the only model that has an average width that corresponds to the amount of shift observed and provides the highest level of average coverage. 

\subsection{CIFAR-10}
\begin{figure}[h]
\centering
\includegraphics[width=.5\textwidth]{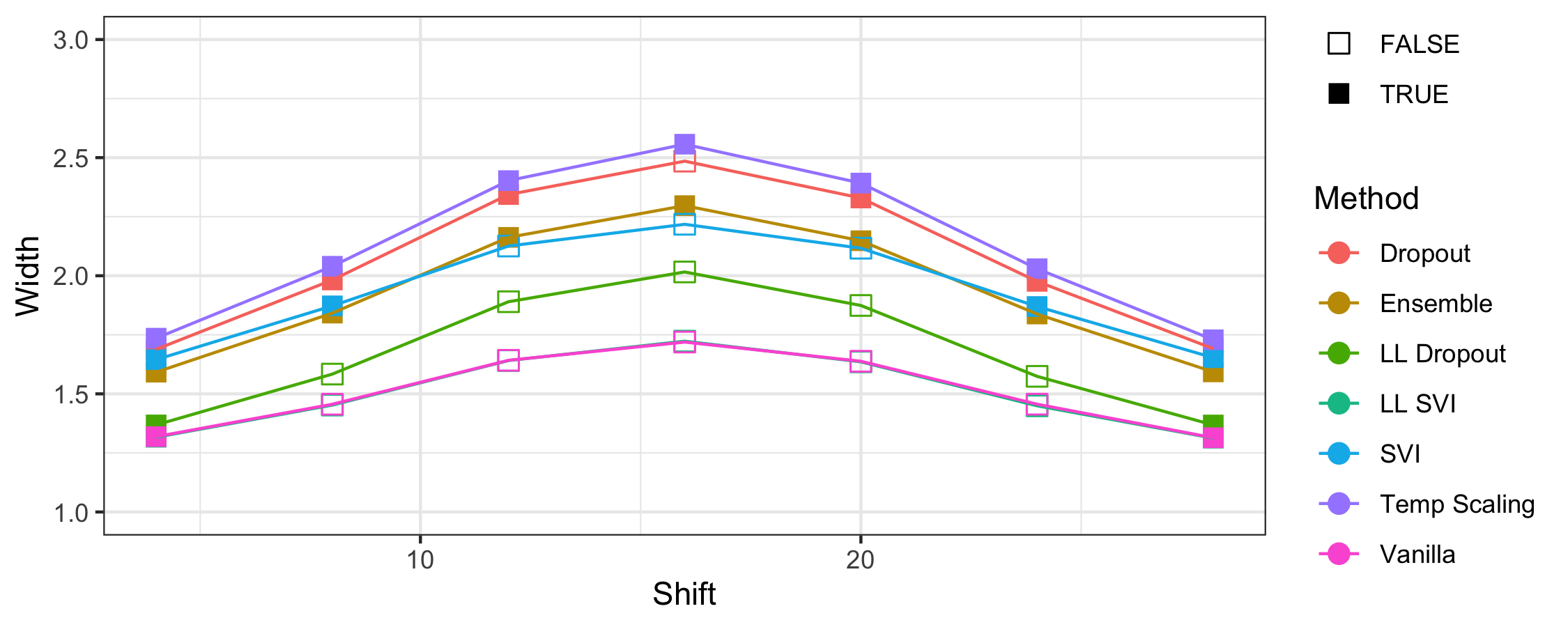}
\caption{The effect of translation shifts on coverage and width in CIFAR-10 images. Coverage remains robust across all pixel shifts, while width increases.}
\label{fig:cifar-translation}
\end{figure}

Next, we consider a more complex image dataset, CIFAR-10. \citet{Ovadia2019-tt} trained 20 layer and 50 layer ResNets. Figure \ref{fig:cifar-translation} shows how the width of the prediction sets increases as translation shift increases. This shift ``rolls'' the image pixel by pixel such that the right most column in the image becomes the left most image. Temperature scaling and ensemble, in particular, have at least 95\% coverage for every translation, although all methods have high levels of coverage on average (though not exceeding 95\%). We find that this high coverage comes with increases in width as shift increases. Figure \ref{fig:cifar-translation} shows that temperature scaling has the highest average width across all models and shifts. Ensembling has the lowest width for the methods that maintain coverage of at least 95\% across all shifts. 

All models have the same encouraging pattern of width increasing as shift increases up to 16 pixels, then decreasing. As CIFAR-10 images are 28 pixels in width and height, this maximum width occurs when the original center of the image is rolled over to and broken by the edge of the image. This likely breaks common features that the methods have learned for classification onto both sides of the image, resulting in decreased classification accuracy and higher levels of uncertainty. 

Between the models which satisfy 95\% coverage levels on all shifts, ensemble models have lower width than temperature scaling models. Under translation shifts on CIFAR-10, ensemble methods perform the best given their high coverage and lower width. 

\begin{figure}[h]
    \centering
    \includegraphics[width=.5\textwidth]{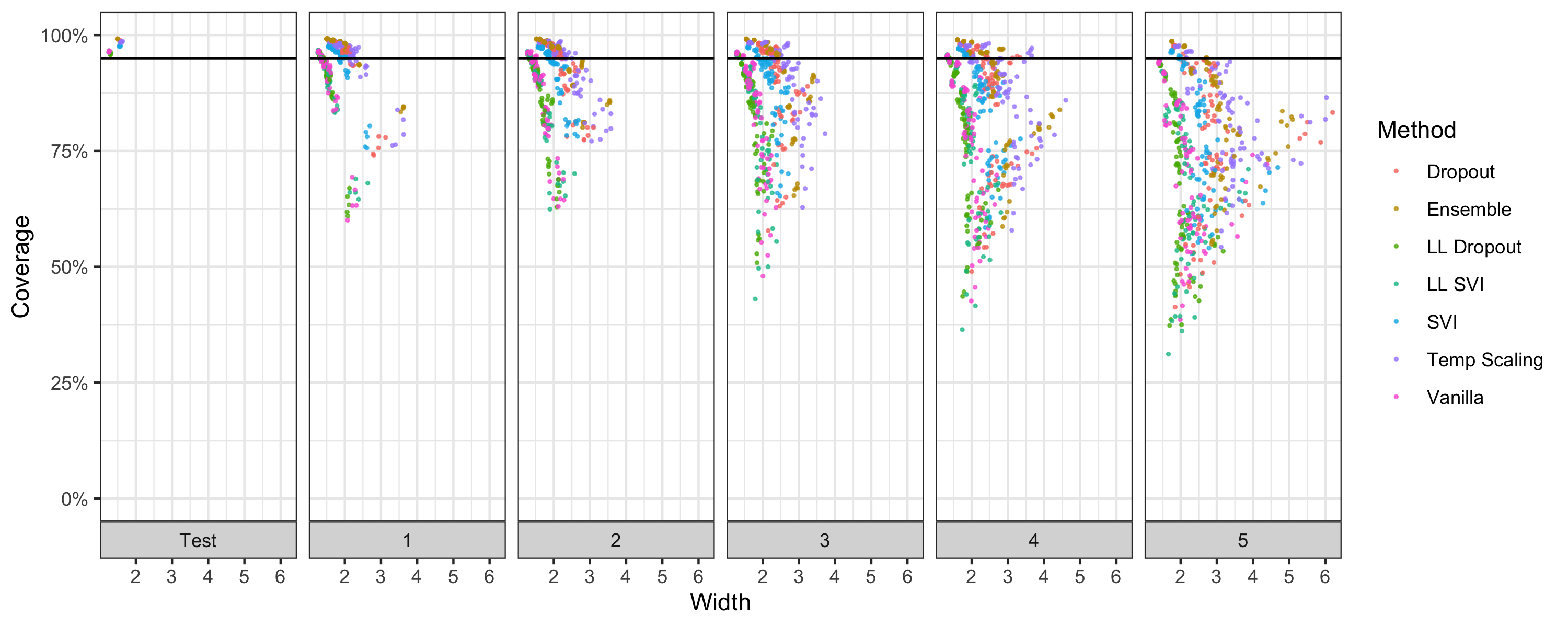}
    \caption{The effect of corruption intensity on coverage levels vs. width in CIFAR-10-C. Each facet panel represents a different corruption level, while points are the coverage of a model on one of 16 corruptions. Each facet has 80 points per method, since 5 iterations were trained per method. For methods with points at the same coverage level, the superior method is to the left as it has a lower width. }
    \label{fig:cifar-coverage_v_width}
\end{figure}

\begin{figure}[h]
    \centering
    \includegraphics[width=.5\textwidth]{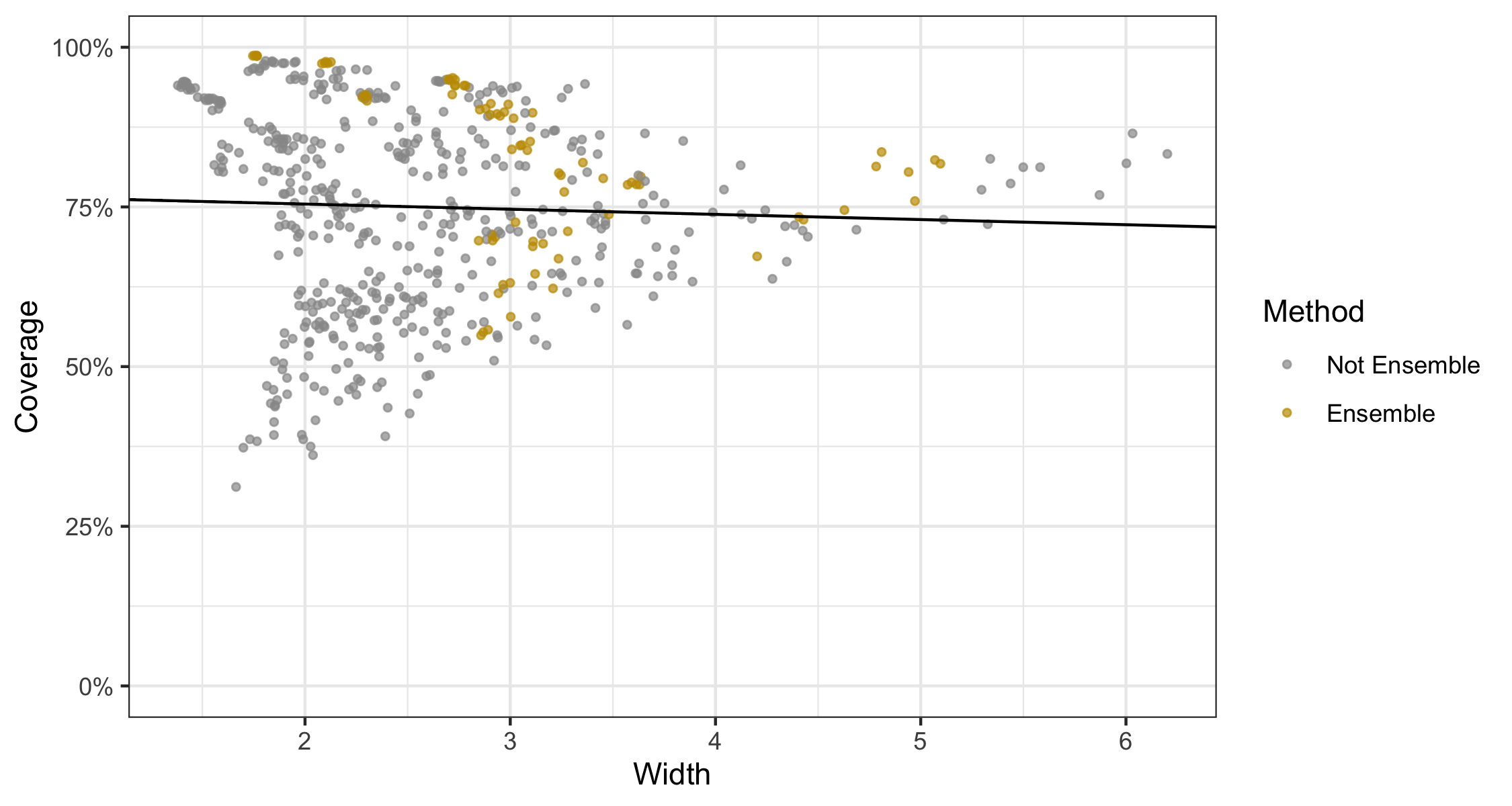}
    \caption{The coverage and width of ensemble and non-ensemble methods at the fifth level out of five levels of corruption in CIFAR-10-C. The black line is a simple linear regression of coverage against width. We then can consider the fraction of points for a particular method (in this case, ensembling) that are above the regression line (see Figures \ref{fig:cifar_percents} and \ref{fig:imagenet_percents}). The higher the fraction of these points above the regression line, the better the method is at providing higher coverage at a relatively smaller width than other methods.}
    \label{fig:cifar_level_3}
\end{figure}

\begin{figure}[h]
    \centering
    \includegraphics[width=.5\textwidth]{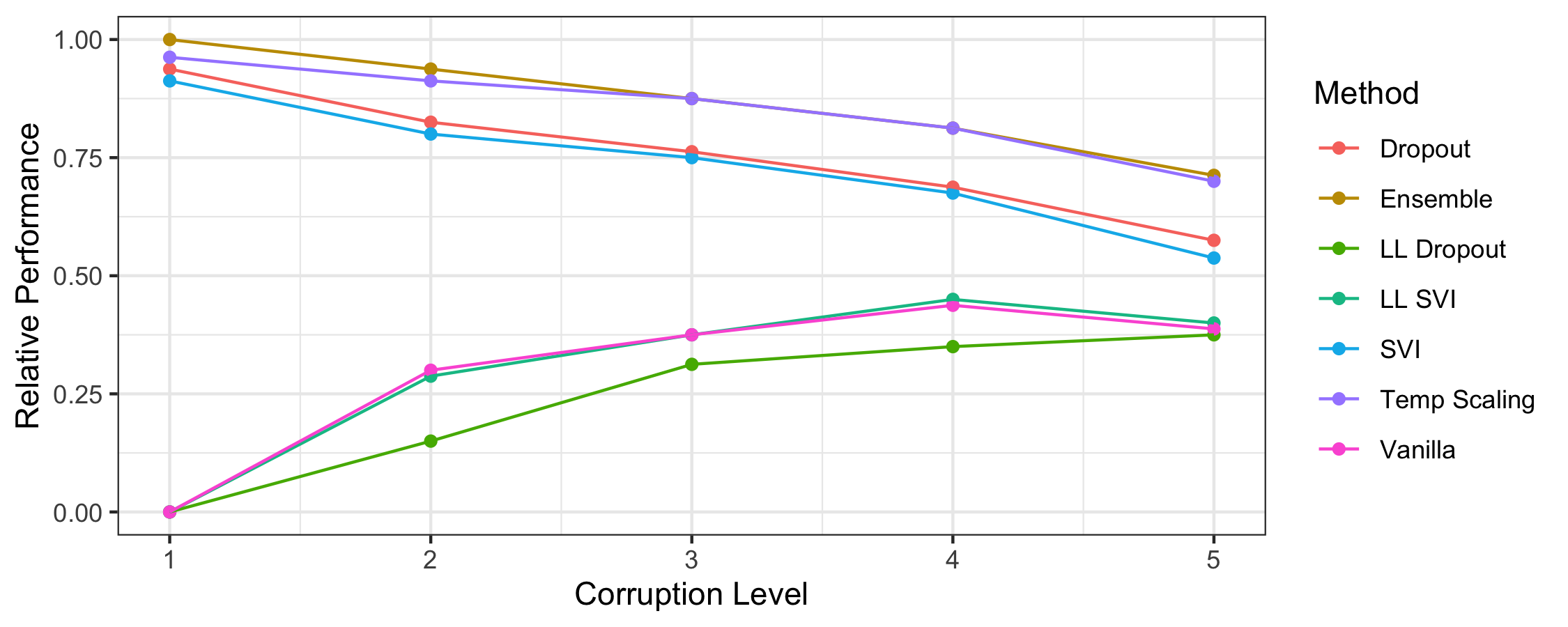}
    \caption{The fraction of marginal coverage levels achieved on CIFAR-10-C corruptions by our assessed methods that are above a regression line of coverage vs width at a specific corruption level. Methods that have better coverage levels at the same width will have a higher fraction of points above the regression line (see Figure \ref{fig:cifar_level_3} for an example). At low levels of shift, dropout, ensemble, SVI, and temperature scaling have strictly better relative performance. As shift increases, poor coverage levels in general cause models to have more parity.}
    \label{fig:cifar_percents}
\end{figure}

\begin{figure}[h]
    \centering
    \includegraphics[width=.5\textwidth]{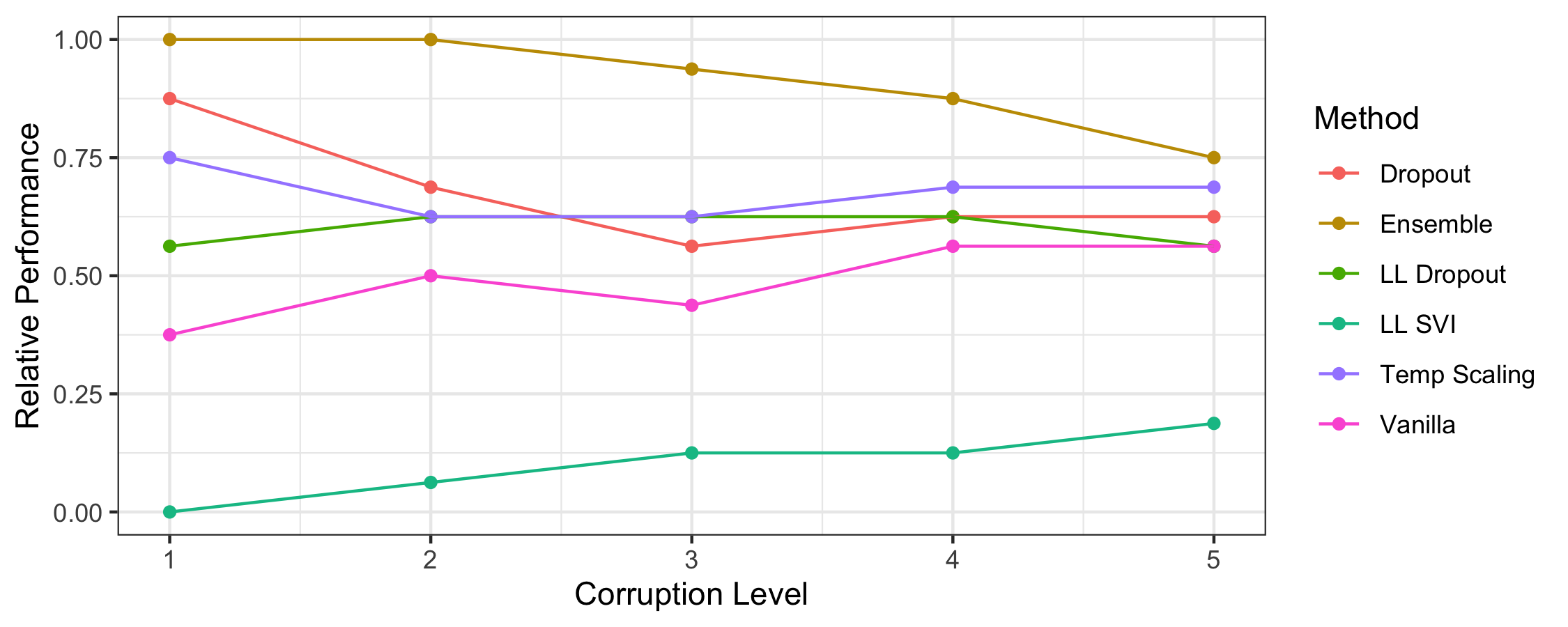}
    \caption{The fraction of marginal coverage levels achieved on ImageNet-C corruptions by our assessed methods that are above a regression line of coverage vs width at a specific corruption level. Methods that have better coverage levels at the same width will have a higher fraction of points above the regression line (see Figure \ref{fig:cifar_level_3} for an example). Ensembling produces the best coverage levels given specific widths across all levels of corruption. However, at higher level of dataset shift, there is more parity between methods.}
    \label{fig:imagenet_percents}
\end{figure}

Additionally, we consider the coverage properties of models on 16 different corruptions of CIFAR-10 from Hendrycks and Gimpel \citep{Hendrycks2016-ov}. Figure \ref{fig:cifar-coverage_v_width} shows coverage vs. width over varying levels of shift intensity. Models that have more dispersed points to the right have higher widths for the same level of coverage. An ideal model would have a cluster of points above the 95\% coverage line and be far to the left portion of each facet. For models that have similar levels of coverage, the superior method will have points further to the left. 

Figure \ref{fig:cifar-coverage_v_width} demonstrates that at the lowest shift intensity, ensemble models, dropout, temperature scaling, and SVI were able to generally provide high levels of coverage on most corruption types. However, as the intensity of the shift increases, coverage decreases. Ensembles and dropout models have for at least half of their 80 model-corruption evaluations at least 95\% coverage up to the third intensity level. At higher levels of shift intensity, ensembles, dropout, and temperature scaling consistently have the highest levels of coverage. Although these higher performing methods have similar levels of coverage, they have different widths. 

\begin{figure*}[h]
    \centering
    \includegraphics[width=\textwidth]{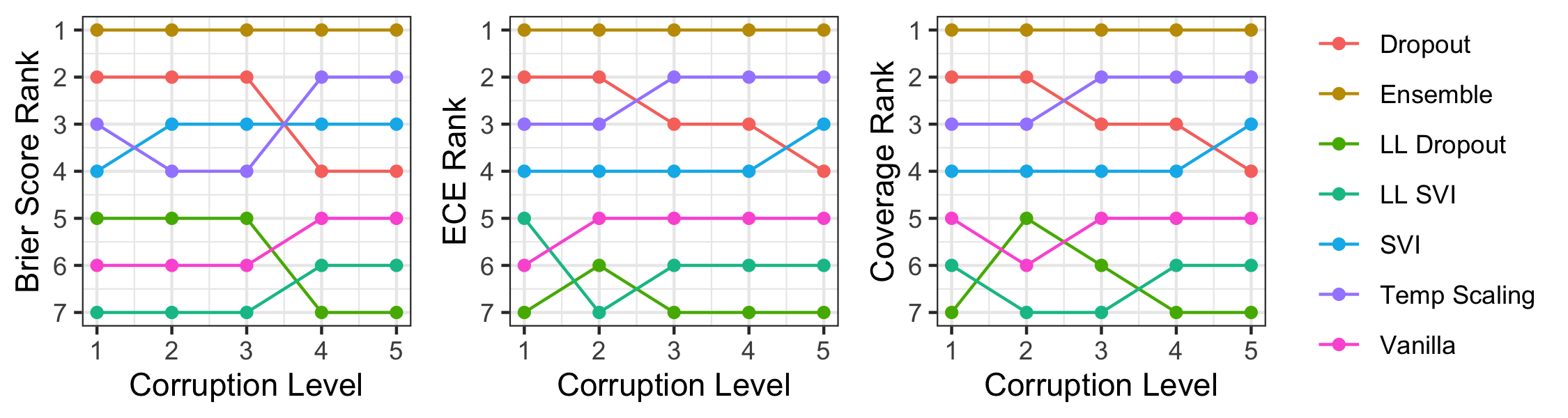}
    \caption{The ranks of each method's performance with respect to each metric we consider on CIFAR-10-C. For Brier Score and ECE, lower is better, while for coverage, higher is better. We observe that all three metrics have a generally consistent ordering, with coverage closely corresponding to the rankings of ECE.}
    \label{fig:rank}
\end{figure*}

We also present a way to quantify the relative strength of each method over a specific level of corruption. In Figure \ref{fig:cifar_level_3}, for instance, we plot only the coverage and widths of methods at the third level of corruption and use the fraction of the points of a particular method that lie above the regression line. Methods that are more effective are providing higher coverage levels at lower widths will have more points above this regression line.

For each of the five corruption levels, we calculated a regression line that modeled coverage as a function of width. Figure \ref{fig:cifar_percents} presents the fraction of marginal coverages on various CIFAR-10-C datasets for each method that exceeded the linear regression prediction. The larger the fraction, the better the marginal coverage of a method given a prediciton interval/set of a particular width. We observe that dropout and ensembles have a strong relative performance to the other methods across all five levels of shift.

Finally, we compared the relative rank order of these methods across coverage as well as two common metrics in uncertainty quantification literature: Brier score and ECE. Figure \ref{fig:rank} shows that the rankings are similar across methods. In particular, coverage has a nearly identical pattern to ECE, with changes only in the lower ranking methods.  


\subsection{ImageNet}

Finally, we analyze coverage and width on ImageNet and ImageNet-C from \citet{Hendrycks2016-ov}. Figure \ref{fig:imagenet-corruption_v_width} shows similar coverage vs. width plots to Figure \ref{fig:cifar-coverage_v_width}. We find that over the 16 different corruptions at 5 levels, ensembles, temperature scaling, and dropout models had consistently higher levels of coverage. Unsurprisingly, Figure \ref{fig:imagenet-corruption_v_width} shows that these methods have correspondingly higher widths. Figure \ref{fig:imagenet_percents} reports the relative performance of each method across corruption levels. Ensembles had the highest fraction of marginal coverages on ImageNet-C datasets above the regression lines at each corruption level. Dropout, LL Dropout, and temperature scaling all had similar performances, while LL SVI a much lower fraction of marginal coverages about the regression lines. None of the methods have a commensurate increase in width to maintain the 95\% coverage levels seen on in-distribution test data as dataset shift increases.

\section{Discussion}
We have provided the first comprehensive empirical study of the frequentist-style coverage properties of popular uncertainty quantification techniques for deep learning models. In regression tasks, LL SVI, SVI, and Gaussian processes all had high levels of coverage across nearly all benchmarks. LL SVI, in particular, had the lowest widths. SVI also had excellent coverage properties across most tasks with tighter intervals than GPs and linear regression. In contrast, the methods based on ensembles and Monte Carlo dropout had significantly worse coverage due to their overly confident and tight prediction intervals. 

In the classification setting, all methods showed very high coverage in the i.i.d setting (i.e. no dataset shift), as coverage is reflective of top-1 accuracy in this scenario. On MNIST data, SVI had the best performance, maintaining high levels of coverage under slight dataset shift and scaling the width of its prediction intervals more appropriately as shift increased relative to other methods. On CIFAR-10 data and ImageNet, ensemble models were superior. They had the highest coverages relative to other methods as demonstrated in Figures \ref{fig:cifar_percents} and \ref{fig:imagenet_percents}. 

An important consideration throughout this work is the choice of hyperparameters in most all of the analyzed methods makes a significant impact on the uncertainty estimates. We set hyperparameters and optimized model parameters according to community best practices in attempt to reflect what a ``real world'' machine learning practitioner might do: selecting hyperparameters based on minimizing validation loss over nested cross validation. Our work is a measurement of the empirical coverage properties of these methods as one would typically utilize them, rather than an exploration of how pathological hyperparameters can skew uncertainty estimates to 0 or to infinity. 

Of particular note is that the width of a prediction interval or set typically correlated with the degree of dataset shift. For instance, when the translation shift is applied to MNIST, both prediction set width and dataset shift is maximized around 14 pixels. There is a 0.9 Pearson correlation between width and shift. Width can serve as a soft proxy of dataset shift and potentially detect shift in real world scenarios. Simultaneously, the ranks of coverage, Brier score, and ECE all are generally consistent. However, coverage is arguably the most interpretable to downstream users of machine learning models. Clinicians, for instance, may not have the technical training  to have an intuition about what specific values of Brier score or ECE mean in practice, while coverage and width are readily understandable. \cite{Manrai2014-xo} already demonstrated clinicians' general lack of intuition about positive predictive value and these uncertainty quantification metrics are more difficult to internalize than PPV.

In summary, we find that popular uncertainty quantification methods for deep learning models do not provide good coverage properties under moderate levels of dataset shift. Although the width of prediction regions do increase under increasing amounts of shift, these changes are not enough to maintain the levels of coverage seen on i.i.d data. We conclude that the methods we evaluated for uncertainty quantification are likely insufficient for use in high-stakes, real-world applications where dataset shift is likely to occur. However, marginal coverage of a prediction interval or set is a natural and intuitive metric to quantify uncertainty. The width of a prediction interval/set is an additional tool that captures dataset shift and provides additional interpretable information to downstream users of machine learning models. 

\bibliography{uai2021-template}

\begin{acknowledgements} 
We would like to thank Alex D'Amour and Balaji Lakshminarayanan for their insightful comments on a draft of this manuscript. 
\end{acknowledgements}

\newlength{\originalVOffset}
 \newlength{\originalHOffset}
 \setlength{\originalVOffset}{\voffset}   
 \setlength{\originalHOffset}{\hoffset}

 \setlength{\voffset}{0cm}
 \setlength{\hoffset}{0cm}
 \AtEndDocument{\includepdf[pages=-]{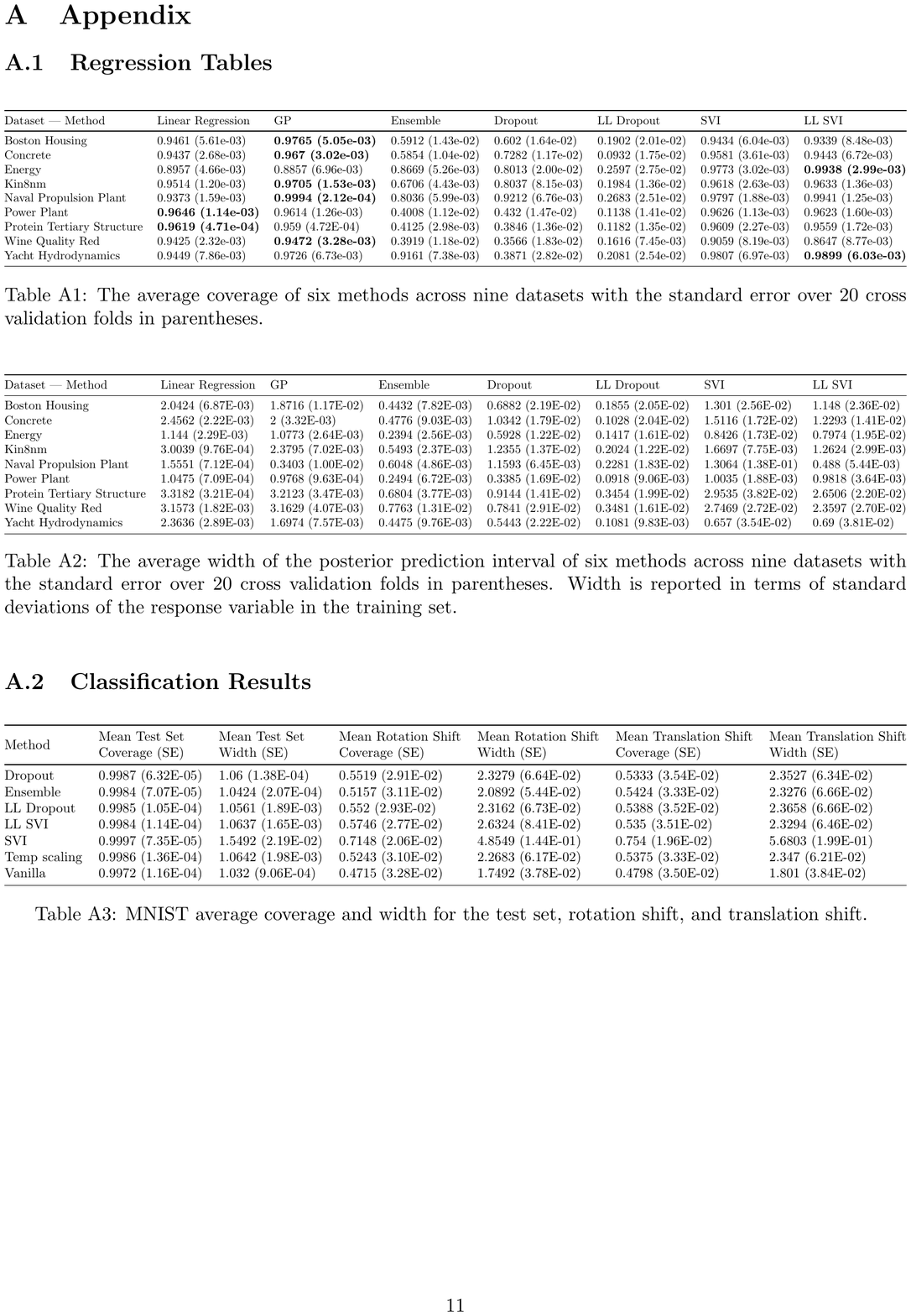}}
 \setlength{\voffset}{\originalVOffset}
 \setlength{\hoffset}{\originalHOffset}
\end{document}